\documentclass[letterpaper, 10 pt, conference]{ieeeconf}  % Comment this line out if you need a4paper

\IEEEoverridecommandlockouts                              % This command is only needed if 
\usepackage{graphics} % for pdf, bitmapped graphics files
\usepackage{epsfig} % for postscript graphics files
\usepackage{caption}
\usepackage{mathrsfs}
\usepackage{booktabs} % 你已使用\toprule，必须加载此包
\usepackage{tabularx} % 实现自动换行的表格核心包
\usepackage{ragged2e} % 优化换行文本的对齐效果

\usepackage[T1]{fontenc} % optional
\usepackage{cite}
\usepackage{booktabs}
\usepackage{amsmath,amssymb,amsfonts}
\usepackage{algorithmic}
\usepackage{setspace}
\usepackage{subfigure}
\usepackage{graphicx}
\usepackage{textcomp}
\usepackage{pifont}
\usepackage[table]{xcolor}
\usepackage{multirow}
\usepackage{threeparttable}
\usepackage{tabularx}
\usepackage{mathtools}
\usepackage[free-standing-units=true]{siunitx}
\usepackage{float}

\newcommand{\rom}[1]{\uppercase\expandafter{\romannumeral #1\relax}}

\usepackage{array}
\newcommand{\PreserveBackslash}[1]{\let\temp=\\#1\let\\=\temp}
\newcolumntype{C}[1]{>{\PreserveBackslash\centering}p{#1}}
\newcolumntype{R}[1]{>{\PreserveBackslash\raggedleft}p{#1}}
\newcolumntype{L}[1]{>{\PreserveBackslash\raggedright}p{#1}}

\definecolor{ourgray}{gray}{0.9}

\makeatletter
\let\NAT@parse\undefined
\makeatother

\usepackage[colorlinks]{hyperref}
  \hypersetup{
    citecolor=blue,
    linkcolor=blue,   
    urlcolor=blue}

\author{Jialong Liu$^{1}$, Dehan Shen$^{1}$, Yanbo Wen$^{1}$, Zeyu Jiang$^{1}$, and Changhao Chen$^{1\dagger}$% <-this % stops a space
\thanks{$^{1}$ PEAK-Lab, The Hong Kong University of Science and Technology (Guangzhou), Guangzhou, 511453, China.}%
\thanks{$\dagger$Corresponding author email: changhaochen@hkust-gz.edu.cn}%
}

\begin{document}

\title{\bf{\LARGE
REAL: Robust Extreme Agility via Spatio-Temporal Policy Learning and Physics-Guided Filtering}\\
% \bf\Large{\href{https://jialonglong.github.io/REAL_wb/}{REAL.github.io}}
}

\maketitle

\begin{abstract}
Extreme legged parkour demands rapid terrain assessment and precise foot placement under highly dynamic conditions. While recent learning-based systems achieve impressive agility, they remain fundamentally fragile to perceptual degradation, where even brief visual noise or latency can cause catastrophic failure. To overcome this, we propose Robust Extreme Agility Learning (REAL), an end-to-end framework for reliable parkour under sensory corruption. Instead of relying on perfectly clean perception, REAL tightly couples vision, proprioceptive history, and temporal memory. We distill a cross-modal teacher policy into a deployable student equipped with a FiLM-modulated Mamba backbone to actively filter visual noise and build short-term terrain memory actively. Furthermore, a physics-guided Bayesian state estimator enforces rigid-body consistency during high-impact maneuvers. Validated on a Unitree Go2 quadruped, REAL successfully traverses extreme obstacles even with a 1-meter visual blind zone, while strictly satisfying real-time control constraints with a bounded 13.1 ms inference time. The code will be released at \href{https://jialonglong.github.io/REAL_wb/}{https://jialonglong.github.io/REAL\_wb/}.
%\textbf{}
% The code and demo is available at\href{https://mac-vo.github.io}{https://mac-vo.github.io}
\end{abstract}

% \begin{keywords}

% \end{keywords}

\section{Introduction}
% ======================================================================
% Paragraph 1: Context (Locomotion -> Parkour)
% ======================================================================
Extreme legged parkour seeks to endow bipedal and quadrupedal robots with the ability to execute highly agile maneuvers across discontinuous, cluttered, and dynamically challenging terrains. Such tasks require rapid terrain assessment, gait transitions, precise foot placement, and continuous balance regulation—all under strict timing and torque constraints. Advancing robotic parkour not only pushes the limits of dynamic locomotion, but also lays the foundation for deploying robots in safety-critical applications such as disaster response, industrial inspection, and planetary exploration in unstructured environments.

Recent progress in reinforcement learning (RL) has largely advanced agile locomotion. Early vision-free policies \cite{hwangbo2019learning, lee2020learning, kumar2021rma} demonstrated impressive dynamic capabilities but lacked environmental awareness. More recent perception-driven systems incorporate privileged terrain information during training and distill high-performance teacher policies into deployable student models \cite{cheng2024extreme, zhuang2023robot}. These sim-to-real pipelines enable robots to operate near their physical limits while reducing manual reward engineering.

Despite recent advances in agile legged locomotion, existing methods remain highly vulnerable to perceptual degradation. Extreme parkour entails impacts, rapid rotations, flight phases, and motion blur, where even brief visual corruption can cause catastrophic failure. Most approaches treat perception as a direct feedforward input to control, without modeling observation uncertainty, exploiting temporal memory, or enforcing physics consistency, allowing visual noise to propagate directly to motor commands and destabilize behavior near dynamic limits.
Moreover, state estimation is often performed by deterministic regressors that neither quantify uncertainty nor respect rigid-body dynamics, making them fragile under slip or high-impact events. Modular mode-switching solutions \cite{zhang2025renet} partially alleviate these issues but introduce additional complexity and heuristic transitions, limiting robustness and scalability.
Collectively, existing research lacks structured cross-modal terrain–body reasoning, uncertainty-aware and physics-consistent state estimation, and stable policy distillation under perceptual corruption.

\begin{figure}[t]
    \centering
    \includegraphics[width=\linewidth]{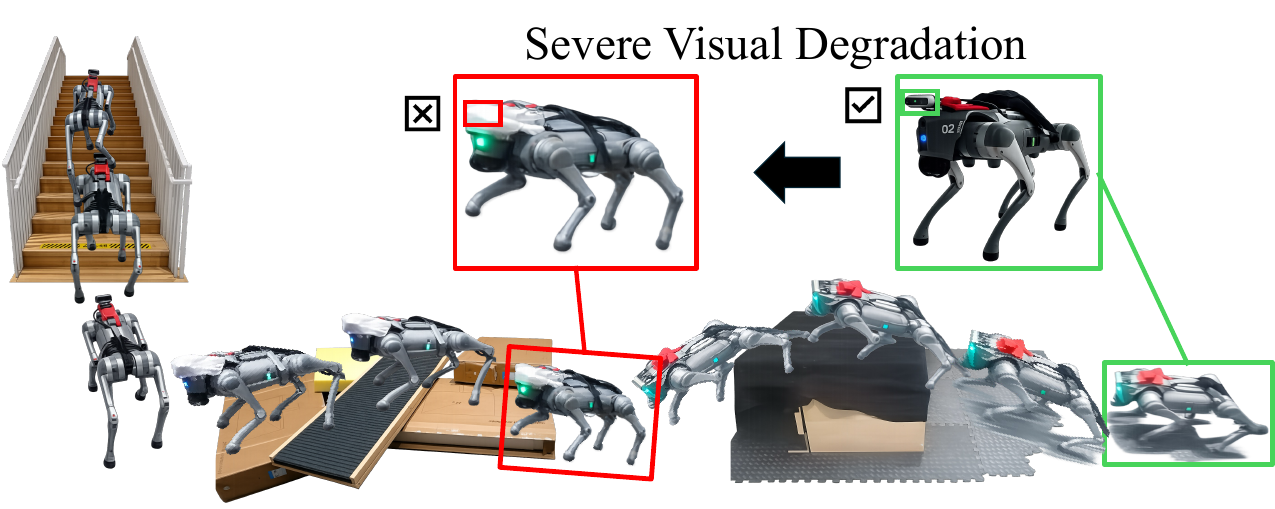}
    \caption{Robust extreme parkour with proposed REAL framework. The robot successfully chains highly dynamic maneuvers across complex terrains with nominal vision (green box), and maintains stable locomotion even under severe visual degradation (red box).}
    \label{fig:teaser}
\end{figure}

To overcome these challenges, we propose \textbf{REAL} (\textbf{R}obust \textbf{E}xtreme \textbf{A}gility \textbf{L}earning), an end-to-end framework for quadrupedal locomotion under perceptual corruption. REAL unifies structured terrain reasoning, adaptive spatio-temporal memory, and physics-guided state estimation within a single training pipeline. A privileged teacher learns precise proprioception–terrain associations through cross-modal attention, which are distilled into a student equipped with FiLM-based cross-modal fusion and a Mamba temporal backbone to suppress unreliable visual features and maintain short-term terrain memory. A physics-guided Bayesian estimator fuses uncertainty-aware neural velocity predictions with rigid-body dynamics via an EKF to ensure consistent motion estimation under impacts and slip. Finally, a consistency-aware loss gating mechanism stabilizes policy distillation and improves sim-to-real transfer.
Extensive simulation and hardware experiments demonstrate zero-shot transfer to a real quadrupedal platform, achieving reliable extreme parkour even under severe perceptual degradation and outperforming prior baselines.

Our contributions are summarized as follows:
\begin{itemize}
\item We propose REAL, a novel spatio-temporal policy learning framework that captures structured cross-modal proprioception–terrain associations, enabling robust and computationally efficient quadrupedal locomotion.
\item We introduce a physics-guided filtering module that explicitly models uncertainty and enforces rigid-body consistency, ensuring reliable state estimation during high-dynamic maneuvers.
\item We develop a consistency-aware loss gating strategy that adaptively balances imitation and reinforcement learning, stabilizing policy distillation and improving sim-to-real robustness.
\item We validate REAL through extensive simulation and hardware experiments, demonstrating zero-shot transfer to a real quadrupedal robot and reliable extreme terrain traversal under severe perceptual degradation.
\end{itemize}

% ======================================================================
% Paragraph 5: Contributions
% ======================================================================

%\item A new distillation framework that uses an Mamba-FiLM architecture to create reliable short-term terrain memory. This enables the system to combine different sensor data, making the policy robust in the event of sensor problems.
    
%\item A physics-based estimator which combines neural predictions with kinematic constraints in order to reduce pose estimation errors. This enables the robot to place its feet accurately and maintain stability during fast or strong movements.
    
%\item We demonstrate zero-shot sim-to-real transfer of our policy on a real quadrupedal robot, enabling it to reliably navigate complex environments. Our method outperforms the baseline even when there is no visual input.

\section{Related Works}

\subsection{Legged Locomotion}
Classical locomotion methods use model-based control to design walking controllers \cite{bellicoso2018dynamic, winkler2015planning, farshidian2017real}. These methods typically involve splitting the task into elevation mapping and foothold planning \cite{agarwal2023egocentric}. 
However, elevation mapping is sensitive to noise and can fail in poor conditions \cite{agarwal2023egocentric, hartley2018legged}. Noisy terrain maps can also cause planning errors. To address this issue, RL-trained controllers have been proposed to adapt to dynamic changes in both quadrupedal and bipedal platforms \cite{siekmann2021blind, radosavovic2023real, li2023robust}. 
To bridge the sim-to-real gap, recent methods have been developed that train policies in simulation with a wide range of randomized physical parameters and deploy them alongside concurrent state estimators \cite{hwangbo2019learning, lee2020learning, kumar2021rma, ji2022concurrent, margolis2022walk}. 
These frameworks use exteroceptive feedback to create latent vectors that describe the local terrain \cite{miki2022learning, yu2021visual, agarwal2023egocentric}. This enables robust locomotion in unstructured environments. However, these methods rely on continuous visual input and may malfunction if the sensors are not working properly. While some approaches incorporate multi-modal delay randomization to alleviate sensor vulnerabilities \cite{imai2022vision}, the fundamental reliance on clean perception remains a bottleneck.

\subsection{Robotic Parkour}
Robotic parkour requires robots to move quickly over obstacles and operate at the limits of their hardware capabilities. Advanced RL methods such as RPL \cite{zhuang2023robot}, Extreme Parkour \cite{cheng2024extreme}, and SoloParkour \cite{chane2024soloparkour}, alongside recent agile navigation frameworks for quadrupeds \cite{hoeller2023anymal} and extreme jumping for bipeds \cite{margolis2021learning, zhao2024humanoid}, have achieved strong results. 
Despite these successes, they often fail when they rely too heavily on continuous visual input. These methods lack the short-term memory required to retain information about recent terrain. Consequently, they are unable to use past observations to address visual issues caused by noise, latency, or temporary sensor signal loss during high-impact dynamic maneuvers.

\subsection{Robust Sequence Modeling and Estimation}
To mitigate the effects of imprecise sensing, previous studies \cite{miki2022learning, rudin2022learning} employed recurrent neural networks (RNNs), while others introduced Cross-Modal Transformers \cite{yang2022learning}, to enable the robot to develop a short-term memory of its own observations. However, noisy or delayed perception can still hinder action planning, and computational delays—especially those inherent to Transformer-based architectures—can disrupt the connection between sensing and control, which is essential for rapid execution. 
For continuous real-time control, state space models such as Mamba \cite{gu2024mamba} outperform standard recurrent and attention-based models by providing bounded $O(1)$ inference complexity \cite{wang2026locomamba}. Building on this, we have developed a system that combines a teacher policy, which is trained using additional privileged information, with a Mamba-FiLM student. This enables the policy to estimate terrain states and robustly recover hidden terrain features using the robot's proprioceptive movement history.
\section{Method}
\label{sec:method}
\begin{figure*}[h]
    \centering
    \vspace{5pt}
    \includegraphics[width=0.95\textwidth]{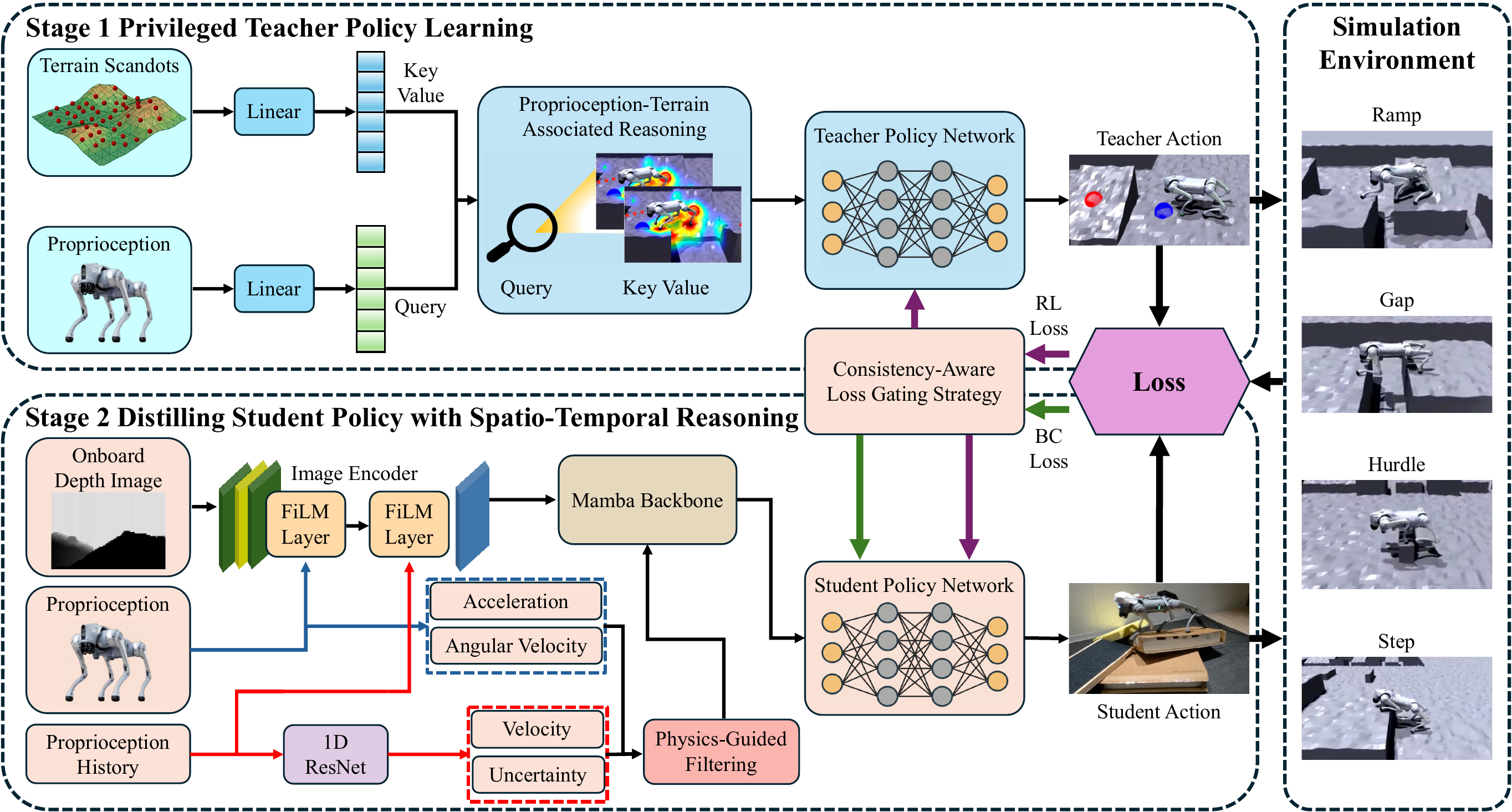}
    \caption{System architecture of REAL. Stage 1(Privileged Teacher Policy Learning) trains a privileged teacher policy via Proprioception-Terrain Associated Reasoning. Stage 2(Distillation Student Policy Learning) distills a deployable student policy using an onboard Mamba-FiLM spatial-temporal backbone and a physics-guided filtering, stabilized by a consistency-aware loss gating strategy.}
    \label{fig:REAL_architecture}
\end{figure*}

%Inputs are first processed via FiLM for cross-modal modulation, then fed with velocity estimates from a physics-guided Filtering into a Mamba backbone for temporal memory. A consistency-aware loss gating strategy dynamically balances imitation and RL to guide the distillation.
%, where all actions from both teacher and student policies receive rewards from the simulation environment.
%1) \textit{Cross-Modal Terrain Reasoning (Teacher Model).}
%A cross-attention-based privileged policy actively retrieves terrain features conditioned on body state, enabling precise foot placement at dynamic limits.
%2) \textit{Temporal Adaptive Perception (Student Model).}
%A FiLM-based visual encoder combined with a Mamba sequence backbone builds short-term terrain memory and mitigates perceptual dropout.
Our proposed REAL is an end-to-end policy learning framework designed for robust quadrupedal parkour. In contrast to prior agile locomotion approaches that often degrade under sensor noise or perceptual inaccuracies, REAL tightly integrates vision, proprioception, and temporal memory within a unified spatio-temporal policy learning architecture. The framework adopts a two-stage training paradigm: In Stage 1, a privileged teacher policy is trained via reinforcement learning (RL) in simulation. The teacher has access to proprioceptive observations, terrain scan points, and additional privileged information, enabling it to learn precise proprioception–terrain associations for agile locomotion; In Stage 2, the teacher policy is distilled into a deployable student policy that operates using only onboard depth sensing and proprioception. This distillation process transfers terrain-aware locomotion capabilities from privileged simulation training to real-world perception-constrained execution. 

The REAL framework consists of three key components: 1) Spatio-Temporal Policy Learning: a privileged teacher learns structured proprioception–terrain associations via cross-modal attention, and its distilled student integrates FiLM-based visual–proprioceptive fusion with a Mamba temporal backbone for robust decision-making under perceptual corruption; 2) Physics-Guided Filtering: an uncertainty-aware neural velocity estimator is fused with rigid-body dynamics through an EKF to ensure physically consistent and noise-robust motion estimation. 3) Consistency-Aware Loss Gating: an adaptive gating mechanism balances behavioral cloning and reinforcement learning based on teacher–student action discrepancy to stabilize training and improve sim-to-real transfer.

\subsection{Spatio-Temporal Policy Learning}
We adopt a teacher–student framework for robust quadrupedal locomotion learning. The privileged teacher policy leverages a cross-modal attention mechanism to establish structured proprioception–terrain reasoning. Specifically, terrain features are retrieved and aggregated conditioned on the robot’s proprioceptive state, enabling context-aware and dynamically adaptive locomotion decisions.
The student policy, designed for real-world deployment, replaces privileged terrain access with onboard visual perception. Spatial visual features are adaptively modulated by proprioceptive signals through Feature-wise Linear Modulation (FiLM), enabling effective cross-modal fusion. A Mamba-based temporal backbone further maintains long-horizon memory, allowing temporally coherent decision-making under partial observability and perceptual corruption.

\noindent\textbf{Privileged Teacher Policy Learning.}
Guided by path waypoints, terrain scan dots, and robot proprioception, the teacher policy acquires extreme parkour skills under a unified reward formulation, leveraging automated curriculum learning and massive parallel simulation \cite{cheng2024extreme}.
In our work, the teacher encodes proprioceptive states into Queries via a linear projection, while terrain scan dots are encoded into Keys and Values. We introduce a cross-modal attention mechanism to explicitly model spatial associations between body states and terrain geometry. This design allows the instantaneous body state to interrogate the terrain representation and selectively retrieve environment features relevant to the next action. The attention operation is defined as:

%\textbf{Teacher Model: Active Terrain Reasoning.}
%During Stage I, the teacher policy has access to terrain scandots $\mathbf{m}_t$ in addition to proprioception $\mathbf{p}_t$. Instead of concatenating features, we use cross-attention where proprioception serves as Query and terrain features serve as Keys and Values:

\begin{equation}
\text{Attention}(\mathbf{Q}, \mathbf{K}, \mathbf{V}) 
= 
\text{softmax}\left(\frac{\mathbf{Q}\mathbf{K}^\top}{\sqrt{d_k}}\right)\mathbf{V}.
\end{equation}

Unlike conventional approaches that rely on feature concatenation followed by MLPs~\cite{cheng2024extreme}, our attention-based design enables state-conditioned terrain reasoning. The robot’s proprioceptive state dynamically determines which terrain regions are most relevant for control. Consequently, the teacher learns structured terrain-to-body associations, such as step-edge localization, height discontinuity awareness, and impact-sensitive landing zone identification. This active retrieval mechanism enhances robustness near dynamic stability limits and reduces catastrophic failures during aggressive maneuvers.

%\textbf{Student Model: Adaptive Spatio-Temporal Policy Learning.}
%The deployable student must operate under noisy, delayed, or missing perception. To handle this, we introduce an Adaptive Spatio-Temporal Student Policy Learning module, which unifies sensory gating and temporal memory.

\noindent\textbf{Distilling Student Policy with Spatio-Temporal Reasoning.} 
During real-world deployment, the student policy no longer has access to privileged terrain information and must instead rely on onboard depth images and proprioception. The student model is designed to be computationally efficient and robust to perceptual degradation. Our student policy model primarily consists of three modules: a module that integrates FiLM and Mamba, an EKF-based Physics-guided filtering, and a consistency-aware gating strategy that informs the learning process.

%We utilize FiLM to dynamically modulate incoming visual features conditioned on the robot's proprioceptive state. Rather than simply fusing modalities, this proprioception-driven mechanism actively scales and shifts the visual representations, guiding the network to attend to the most critical terrain features relevant to the robot's current posture. Concurrently, an EKF fuses IMU acceleration measurements with a Bayesian velocity distribution estimated by a ResNet module, yielding a reliable estimate of the robot's base velocity. The FiLM-modulated visual features are then directly concatenated with both the raw proprioceptive data and the EKF-estimated velocities. This comprehensive state vector is subsequently fed into the Mamba backbone, which endows the deployable student policy with robust spatio-temporal memory and reasoning capabilities. Finally, a consistency-aware gating strategy adaptively governs the distillation process, enabling the student to optimally balance privileged teacher imitation with autonomous exploration.

%\subsubsection{Temporal Adaptive Perception}

%Temporal Adaptive Perception module employs FiLM for sensory gating to dynamically adjust the focus on visual information, and utilizes Mamba for temporal memory to process sequential information.

%\textbf{Feature-wise Linear Modulation.}
%\noindent\textbf{FiLM-Based Visual Encoder.} 
FiLM and Mamba form the core of our student's spatio-temporal  processing. Depth images and proprioceptive measurements are first processed through FiLM, where proprioception dynamically modulates visual features to extract terrain representations relevant to the current body state. %Depth images and proprioception are first fed into FiLM, where proprioceptive information dynamically modulates visual features to extract spatial visual features relevant to the current body state. Concurrently, the FiLM-modulated visual features are then concatenated with raw proprioception and velocity estimated from physics-guided filtering, forming a comprehensive state vector that is passed to the Mamba backbone for temporal memory.
%Visual features $\mathbf{F}_{\text{CNN}}$ extracted from depth images after Convolutional Neural Network (CNN) are modulated by proprioception using FiLM:
Let $\mathbf{F}_{\text{CNN}}$ denote visual features extracted by a convolutional neural network (CNN). FiLM modulation is defined as:
\begin{equation}
    \text{FiLM}(\mathbf{F}_{\text{CNN}}) = \boldsymbol{\gamma}(\mathbf{p}_t) \odot \mathbf{F}_{\text{CNN}} + \boldsymbol{\beta}(\mathbf{p}_t),
\end{equation}

%where $\gamma$ and $\beta$ are scaling and shifting parameters derived from proprioceptive information $\textbf{p}_t$. These parameters modulate visual features in a channel-wise manner, ultimately yielding enhanced features that integrate visual and proprioceptive information.
where $\boldsymbol{\gamma}(\mathbf{p}_t)$ and $\boldsymbol{\beta}(\mathbf{p}_t)$ are channel-wise scaling and shifting parameters predicted from the proprioceptive state $\mathbf{p}_t$, and $\odot$ denotes element-wise multiplication. This modulation adaptively gates visual features based on the robot’s instantaneous dynamics.
FiLM serves as an adaptive sensory filter. During impacts or rapid body rotations, unreliable visual features—caused by motion blur or occlusion—are suppressed using fast proprioceptive feedback, thereby reducing erroneous actuation.

The FiLM-modulated visual features are concatenated with raw proprioception and velocity estimates from the physics-guided filter, forming a comprehensive state representation $\mathbf{x}_t$ that is fed into a Mamba backbone for temporal modeling. When exteroceptive input degrades, the controller relies more heavily on temporal memory. Mamba provides efficient sequence modeling with linear-time complexity, making it suitable for onboard deployment.

Given input features $\mathbf{x}_t$, the hidden state $\mathbf{h}_t$ evolves as:
\vspace{-5pt}
\begin{equation}
    \mathbf{h}_t = {\mathbf{A}}_t \mathbf{h}_{t-1} + {\mathbf{B}}_t \mathbf{x}_t, \quad \mathbf{y}_t = \mathbf{C}_t \mathbf{h}_t.
\end{equation}
where $\mathbf{h}_t$ encodes spatio-temporal memory and $\mathbf{y}_t$ is the output. Data-dependent matrices ($\mathbf{A}_t$, $\mathbf{B}_t$, $\mathbf{C}_t$) adapt via $\mathbf{x}_t$ to selectively filter noise and retain terrain features. With RoPE preserving spatial order, this enables real-time terrain reconstruction as a fallback during visual degradation.
%High-dynamic robot parkour violates the no-slip assumptions commonly used in legged odometry. Pure MLP-based velocity regression fails to model uncertainty and temporal consistency. We introduce an Uncertainty-Aware Velocity Network, which combines probabilistic neural velocity estimation an EKF.
\subsection{Physics-Guided Filtering}
High-dynamic quadrupedal parkour frequently violates the no-slip assumptions commonly adopted in legged odometry. Pure MLP-based velocity regression lacks uncertainty modeling and temporal consistency, leading to drift and physically inconsistent estimates. To address this, REAL integrates an uncertainty-aware neural velocity predictor with rigid-body dynamics through an Extended Kalman Filter (EKF). The learned predictor provides adaptive uncertainty estimates, while the dynamics model enforces physical constraints. This hybrid design enhances robustness to perceptual noise and prevents physically implausible state estimates during aggressive maneuvers.

\noindent\textbf{Uncertainty-Aware Velocity Network.}
Existing parkour baselines typically estimate base linear velocity from single-frame observations \cite{cheng2024extreme}, neglecting temporal interaction patterns between the robot and terrain. Instead, we employ a 1D ResNet to extract motion features from a 10-frame proprioceptive sequence, including IMU measurements and joint encoder readings \cite{zhang2025renet}.

The network predicts a Gaussian distribution \(\mathcal{N}(\boldsymbol{v}_{net}, \boldsymbol{\Sigma}_t)\), where $\boldsymbol{v}_{net}$ is the estimated velocity and \(\boldsymbol{\Sigma}_t \in \mathbb{R}^{3 \times 3}\) represents the associated uncertainty. The loss function $\mathcal{L}_{\text{est}}$ adopts a Huber-Gaussian Loss \cite{AirIO}, which integrates the robustness of Huber loss $\mathcal{L}_{huber}$ with probabilistic calibration. Its formulation is given by:

%The network predicts a Gaussian distribution $\mathcal{N}(\mathbf{v}_{\text{net}}, \mathbf{\Sigma}_t)$ with a diagonal covariance matrix $\mathbf{\Sigma}_t = \text{diag}(\sigma_{t,x}^2, \sigma_{t,y}^2, \sigma_{t,z}^2)$, assuming independence between axes. Training is performed using a Huber-Gaussian loss function\cite{AirIO} $\mathcal{L}_{\text{est}}$ that combines robust regression with probabilistic calibration:

%\begin{equation}
%\mathcal{L}_{\text{est}} = \mathcal{L}_{\text{Huber}}(\mathbf{v}_{\text{gt}}, \mathbf{v}_{\text{net}}) + \lambda \sum_{i \in \{x,y,z\}} \left( \frac{(v_{\text{gt}, i} - v_{\text{net}, i})^2}{2\sigma_{t,i}^2} + \ln \sigma_{t,i} \right)
%\end{equation}
%where $\mathbf{v}_{\text{gt}}$ is the ground-truth velocity. $\mathbf{v}_{\text{net}}$ denotes the base linear velocity predicted by ResNet from proprioceptive history. $\lambda = 1 \times 10^{-4}$ balances the two objectives. The Huber function ensures robust mean estimation by mitigating the impact of outliers and is defined as

%where $\delta = 5 \times 10^{-3}$ controls the transition between quadratic and linear loss regions.
\vspace{-5pt}
\begin{equation}
\mathcal{L}_{est} = \mathcal{L}_{huber}(\boldsymbol{v}_{gt}, \boldsymbol{v}_{net}) + \lambda \sum_t \left( \frac{\|\boldsymbol{v}_{gt} - \boldsymbol{v}_{net}\|^2}{2\boldsymbol{\Sigma}_t} + \frac{1}{2} \ln |\boldsymbol{\Sigma}_t| \right)
\end{equation}

\begin{equation}
\mathcal{L}_{Huber} = \sum_t \begin{cases} 
\frac{1}{2} (\boldsymbol{v}_{gt} - \boldsymbol{v}_{net})^2, & \text{if } |\boldsymbol{v}_{gt} - \boldsymbol{v}_{net}| < \delta \\
\delta (|\boldsymbol{v}_{gt} - \boldsymbol{v}_{net}| - \frac{1}{2}\delta), & \text{otherwise}
\end{cases}
\end{equation}
where \(\boldsymbol{v}_{gt}\) denotes the ground-truth velocity, $\delta$ is the threshold. 
The Huber term $\mathcal{L}_{huber}$ ensures robust mean estimation under outliers, while the Gaussian negative log-likelihood term calibrates predictive uncertainty. Consequently, the estimator jointly predicts velocity and confidence. During highly dynamic events—such as flight phases or foot slippage—the network increases its predicted uncertainty, thereby reducing its influence during subsequent fusion \cite{lee2026attention}.

%The first term is Huber loss $\mathcal{L}_{\text{huber}}$, which ensures robust mean estimation by mitigating the impact of outliers. The second term enables the model to calibrate its uncertainty. Through this design, the estimator enables the estimator to jointly predict velocity and confidence. During highly dynamic events such as flight phases or foot slippage, the network increases its predicted uncertainty, thereby reducing its influence during subsequent fusion~\cite{lee2026attention}. 

%The estimator predicts both the speed and its own level of confidence in the result. For instance, if the robot jumps or slips, the estimator recognises that its prediction may be less accurate and reduces its confidence accordingly. In this case, the system relies less on this uncertain prediction and pays more attention to other data sources \cite{lee2026attention}.

\noindent\textbf{Bayesian Fusion via Extended Kalman Filter.}
To enforce physical consistency and mitigate integration drift, we fuse the neural estimate $\boldsymbol{v}_{net}$ with a rigid-body motion model using an EKF~\cite{hartley2018legged}. For computational efficiency and numerical stability, we adopt a diagonal covariance formulation.

\emph{Prediction step.} The velocity is propagated in the body frame using IMU linear acceleration $\mathbf{a}_t$ and angular velocity $\mathbf{\boldsymbol{\omega}}_t$. We assume the error dynamics are dominated by process noise, simplifying the covariance propagation:
\begin{equation}
    \boldsymbol{v}_{t|t-1} = \boldsymbol{v}_{t-1} + (\mathbf{a}_t - \mathbf{\boldsymbol{\omega}}_t \times \boldsymbol{v}_{t-1}) \Delta t
\end{equation}
\begin{equation}
    \mathbf{P}_{t|t-1} = \mathbf{P}_{t-1} + \mathbf{F}
\end{equation}
where $\Delta t$ is the control interval. $\mathbf{F} \in \mathbb{R}^{3 \times 3}$ is the process noise covariance matrix, modeling the drift and noise inherent in IMU integration. In our implementation, $\mathbf{F}$ is set to a constant diagonal matrix. $\mathbf{P}_t \in \mathbb{R}^{3 \times 3}$ represents the state estimation covariance, tracking the accumulated uncertainty of the fused velocity $\boldsymbol{v}_t$.

\emph{Update step.} The neural prediction is treated as a virtual measurement. The Kalman gain $\mathbf{E}$ adapts automatically according to the predicted covariance:
\begin{equation}
    \mathbf{E}_t = \mathbf{P}_{t|t-1} (\mathbf{P}_{t|t-1} + \boldsymbol{\Sigma}_t)^{-1}
\end{equation}
\begin{equation}
    \boldsymbol{v}_t = \boldsymbol{v}_{t|t-1} + \mathbf{E}_t (\boldsymbol{v}_{net} - \boldsymbol{v}_{t|t-1})
\end{equation}
\begin{equation}
    \mathbf{P}_t = (\mathbf{I} - \mathbf{E}_t)\mathbf{P}_{t|t-1}
\end{equation}
When the neural predictor reports high uncertainty, the Kalman gain decreases, reducing correction magnitude. Conversely, confident predictions exert stronger influence. This probabilistic coupling between learning-based estimation and physics-based prediction ensures stable and drift-resistant velocity tracking under impacts, slippage, and partial sensor degradation.
\begin{table*}[t]
    \centering
    \vspace{-2pt}
    \caption{Quantitative Performance Breakdown Across Diverse Parkour Terrains}
    \label{tab:extreme_terrain}
    
    \resizebox{\textwidth}{!}{%
    \begin{tabular}{l cc cc cc ccccc}
        \toprule
        
        \multirow{2}{*}{\textbf{Method}} & 
        \multicolumn{2}{c}{\textbf{Hurdles}} & 
        \multicolumn{2}{c}{\textbf{Steps}} & 
        \multicolumn{2}{c}{\textbf{Gaps}} & 
        \multicolumn{5}{c}{\textbf{Overall Average}} \\
        
        \cmidrule(lr){2-3} \cmidrule(lr){4-5} \cmidrule(lr){6-7} \cmidrule(lr){8-12}
        
        & \textbf{SR} ($\uparrow$) & \textbf{MXD} ($\uparrow$) 
        & \textbf{SR} ($\uparrow$) & \textbf{MXD} ($\uparrow$) 
        & \textbf{SR} ($\uparrow$) & \textbf{MXD} ($\uparrow$) 
        & \textbf{SR} ($\uparrow$) & \textbf{MXD} ($\uparrow$) & \textbf{MEV} ($\downarrow$) & \textbf{Time} ($\downarrow$) & \textbf{Coll.} ($\downarrow$) \\
        
        \midrule
        
        Extreme Parkour \cite{cheng2024extreme} & 0.18 & 0.15 & 0.14 & 0.18 & 0.10 & 0.30 & 0.16 & 0.21 & 34.24 & 0.14 & 0.27 \\
        RPL \cite{zhuang2023robot}             & 0.05 & 0.12 & 0.04 & 0.09 & 0.03 & 0.10 & 0.04 & 0.10 & 1.56  & 0.02 & 0.06 \\
        SoloParkour \cite{chane2024soloparkour}     & 0.42 & 0.24 & 0.49 & 0.28 & \textbf{0.36} & 0.34 & 0.39 & 0.34 & 96.93 & 0.06 & 0.09 \\
        
        \midrule
        
        \textbf{REAL (Ours)} & \textbf{0.82} & \textbf{0.42} & \textbf{0.94} & \textbf{0.53} & 0.28 & \textbf{0.39} & \textbf{0.78} & \textbf{0.45} & \textbf{18.41} & \textbf{0.02} & \textbf{0.06} \\
        
        \bottomrule
    \end{tabular}%
    } % 魔法结束符号
    
    \vspace{1pt}
    % 将底部注释严格限制在页面总宽度内
    \begin{minipage}{\textwidth}
        \footnotesize 
        \textit{Note:} We evaluate agility ($\uparrow$) and safety ($\downarrow$). \textbf{SR:} Success Rate—how often the robot reaches all target goals. \textbf{MXD:} Mean X-Displacement $\in [0, 1]$, showing the normalized forward progress. \textbf{MEV:} Mean Edge Violations—the average number of unsafe foot-edge contacts in each episode. \textbf{Time / Coll.:} Timeout rate per episode and torso collision rate per step.
    \end{minipage}
\end{table*}
% --- 大表 2：感知鲁棒性退化 (置于页底，需导言区包含 \usepackage{stfloats}) ---
\begin{table*}[t]
    \centering
    \caption{Robustness Evaluation: Performance Degradation Under Severe Perceptual Interference}
    \label{tab:robustness_drop}
    \resizebox{\textwidth}{!}{%
    \begin{tabular}{l cc cc cc cc}
        \toprule
        
        \multirow{2}{*}{\textbf{Method}} & 
        \multicolumn{2}{c}{\textbf{Nominal (No Interference)}} & 
        \multicolumn{2}{c}{\textbf{Condition A: Frame Drop}} & 
        \multicolumn{2}{c}{\textbf{Condition B: Gaussian Noise}} & 
        \multicolumn{2}{c}{\textbf{Condition C: Spatial FoV Occlusion}} \\
        
        \cmidrule(lr){2-3} \cmidrule(lr){4-5} \cmidrule(lr){6-7} \cmidrule(lr){8-9}
        
        & \textbf{SR ($\uparrow$)} & \textbf{MXD ($\uparrow$)} 
        & \textbf{SR ($\uparrow$)} & \textbf{MXD ($\uparrow$)} 
        & \textbf{SR ($\uparrow$)} & \textbf{MXD ($\uparrow$)} 
        & \textbf{SR ($\uparrow$)} & \textbf{MXD ($\uparrow$)} \\
        
        \midrule
        
        Extreme Parkour \cite{cheng2024extreme} & 0.16 & 0.22 & 0.16 \scriptsize{($\downarrow$ 0.00)} & 0.22 \scriptsize{($\downarrow$ 0.00)} & 0.11 \scriptsize{($\downarrow$ 0.05)} & 0.21 \scriptsize{($\downarrow$ 0.01)} & 0.13 \scriptsize{($\downarrow$ 0.03)} & 0.21 \scriptsize{($\downarrow$ 0.01)} \\
        RPL \cite{zhuang2023robot}              & 0.04 & 0.10 & 0.01 \scriptsize{($\downarrow$ 0.04)} & 0.04 \scriptsize{($\downarrow$ 0.06)} & 0.01 \scriptsize{($\downarrow$ 0.03)} & 0.07 \scriptsize{($\downarrow$ 0.04)} & 0.01 \scriptsize{($\downarrow$ 0.03)} & 0.07 \scriptsize{($\downarrow$ 0.03)} \\
        SoloParkour \cite{chane2024soloparkour}      & 0.39 & 0.35 & 0.20 \scriptsize{($\downarrow$ \textbf{0.19})} & 0.24 \scriptsize{($\downarrow$ 0.11)} & 0.37 \scriptsize{($\downarrow$ 0.03)} & 0.35 \scriptsize{($\downarrow$ 0.00)} & 0.41 \scriptsize{($\uparrow$ 0.02)} & 0.35 \scriptsize{($\uparrow$ 0.00)} \\
        
        \midrule
        
        \textbf{REAL (Ours)} & \textbf{0.78} & \textbf{0.46} & \textbf{0.61} \scriptsize{($\downarrow$ 0.17)} & \textbf{0.41} \scriptsize{($\downarrow$ 0.05)} & \textbf{0.51} \scriptsize{($\downarrow$ 0.27)} & \textbf{0.38} \scriptsize{($\downarrow$ 0.08)} & \textbf{0.72} \scriptsize{($\downarrow$ \textbf{0.06})} & \textbf{0.43} \scriptsize{($\downarrow$ \textbf{0.02})} \\
        
        \bottomrule
    \end{tabular}%
    }
    
    \vspace{1pt}
    \begin{minipage}{\textwidth}
        \footnotesize 
        \textit{Note:} Values in parentheses denote the absolute performance drop ($\downarrow$) or gain ($\uparrow$) compared to the nominal condition (No Interference).
    \end{minipage}
\end{table*}
\subsection{Consistency-Aware Loss Gating Strategy}
%We introduce a consistency-aware loss gating mechanism that dynamically balances behavioral cloning and RL objectives during student training. The weighting is adaptively adjusted according to the discrepancy between teacher and student actions. When the student closely matches the teacher, RL exploration is encouraged; when discrepancies are large, imitation is emphasized. This adaptive transition stabilizes training and significantly improves sim-to-real transfer performance.

To mitigate the sim-to-real gap in Stage II training, we introduce a consistency-aware loss gating strategy that dynamically balances behavioral cloning (BC) and reinforcement learning (RL). The key insight is that the discrepancy between the student policy $\pi_S$ and the privileged teacher policy $\pi_T$ reflects the student’s reliability under perceptual corruption. While the teacher operates with privileged information, the student must act under noisy onboard observations.

\noindent\textbf{Adaptive Loss Gating.}
%We use a gating factor $\lambda$ (between 0 and 1) to adjust how much focus is given to reinforcement learning loss. ($\mathcal{L}_{\text{RL}}$) or Behavioral Cloning loss ($\mathcal{L}_{\text{BC}}$). The value of $\lambda$ depends on how different the actions of the student and teacher are, as measured by their Euclidean distance. We then convert this distance into a value of $\lambda$ using a Sigmoid function:
We introduce a gating coefficient $\lambda \in [0,1]$ to dynamically balance the RL loss $\mathcal{L}_{\text{RL}}$ and the behavioral cloning loss $\mathcal{L}_{\text{BC}}$. The gating factor is determined by the Euclidean distance between student action $\mathbf{a}_S$ and teacher action $\mathbf{a}_T$:
\begin{equation}
    \lambda = \sigma \left( k \cdot (\tau - \|\mathbf{a}_S - \mathbf{a}_T\|_2) \right),
\end{equation}
%Here, $\sigma(\cdot)$ is the Sigmoid function, $\tau$ is the consistency threshold, and $k$ controls how quickly the change happens \cite{margolis2023walk}. The final goal, $\mathcal{L}_{\text{total}}$, is calculated by combining different parts with certain weights:
where $\sigma(\cdot)$ denotes the sigmoid function, $\tau$ is a consistency threshold, and $k$ controls the sharpness of the transition~\cite{margolis2023walk}. The overall training objective is then defined as:

\begin{equation}
    \mathcal{L}_{\text{total}} = \lambda \mathcal{L}_{\text{RL}} + (1 - \lambda) \mathcal{L}_{\text{BC}}.
\end{equation}

This formulation enables smooth interpolation between imitation-driven stabilization and reinforcement-driven adaptation.

\noindent\textbf{Interpretation.}
The proposed mechanism can be viewed as an adaptive supervision schedule \cite{ross2011reduction}, where the training regime shifts according to the student’s consistency with the teacher:
\begin{equation}
    \mathcal{M}_{\text{train}} \approx \begin{cases}
        \underbrace{\mathcal{L}_{\text{BC}} \quad (\lambda \to 0)}_{\text{Stability Preservation}}, & \text{if } \|\mathbf{a}_S - \mathbf{a}_T\|_2 \gg \tau \\[10pt]
        \underbrace{\mathcal{L}_{\text{RL}} \quad (\lambda \to 1)}_{\text{Robustness Adaptation}}, & \text{otherwise}
    \end{cases}.
\end{equation}

When the action discrepancy is large, imitation learning dominates to prevent destabilizing deviations and ensure safe policy refinement. As the student’s behavior aligns with the teacher, the objective gradually shifts toward RL, encouraging exploration and robustness to perceptual noise. This adaptive transition stabilizes training while promoting resilience, resulting in improved sim-to-real transfer without sacrificing policy performance.

%\noindent If the student's actions differ greatly from those of the teacher ($\|\mathbf{a}_S - \mathbf{a}_T\|_2 \gg \tau$), the system provides strict guidance to prevent the student from making significant errors. However, when the student's actions resemble those of the teacher more closely, the system allows the student to explore independently. This helps the student to improve their ability to handle sensor noise.
\section{Experiment}
\label{sec:experiment}

\begin{figure}[htbp]
    \centering
    \includegraphics[width=\columnwidth]{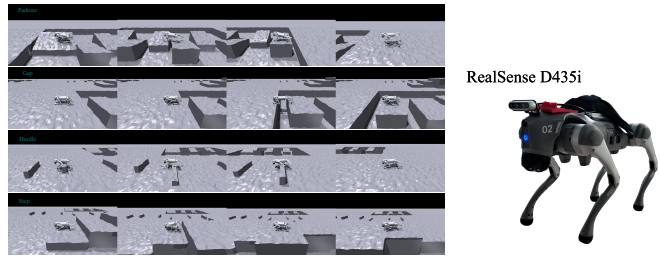}
    \caption{\textbf{Left.} Snapshots of the REAL policy executing dynamic manoeuvres across extreme terrains. \textbf{Right.} Physical hardware setup. The REAL policy has been deployed on a Unitree Go2 quadrupedal robot.}
    \label{fig:hardware_fig}
\end{figure}

We evaluate REAL in Isaac Gym \cite{makoviychuk2021isaac} and on a real robot. Our evaluation focuses on: 1) extreme terrain traversability, 2) robustness to sensory degradation, 3) blind-zone maneuvers, and 4) zero-shot sim-to-real transfer.
% --- A. 实验设置 ---
\subsection{Experimental Setup}
\noindent\textbf{Simulation \& Training.} We train our policy in a simulated environment that automatically generates challenging terrains, including gaps, hurdles, and slanted ramps (Fig. \ref{fig:hardware_fig}). The terrain difficulty scales dynamically according to the robot's proficiency during training. The control policy operates at 50 Hz, outputting target joint angles that are tracked by a low-level PD controller running at 1 kHz \cite{KatzCK19}. The entire policy is trained from scratch in approximately 30 hours using a single NVIDIA RTX 4080 GPU.

\noindent\textbf{Hardware Platform.} We deploy our trained policy on a Unitree Go2 quadrupedal robot (Fig. \ref{fig:hardware_fig}). Proprioceptive data (joint positions, velocities, and IMU) is collected via onboard motor controllers, while exteroception is provided by a front-facing Intel RealSense D435i depth camera. All state estimation and neural network inferences run entirely onboard using an NVIDIA Jetson module.

\begin{figure*}[h]
    \centering
    \vspace{5pt}
    \includegraphics[width=\textwidth]{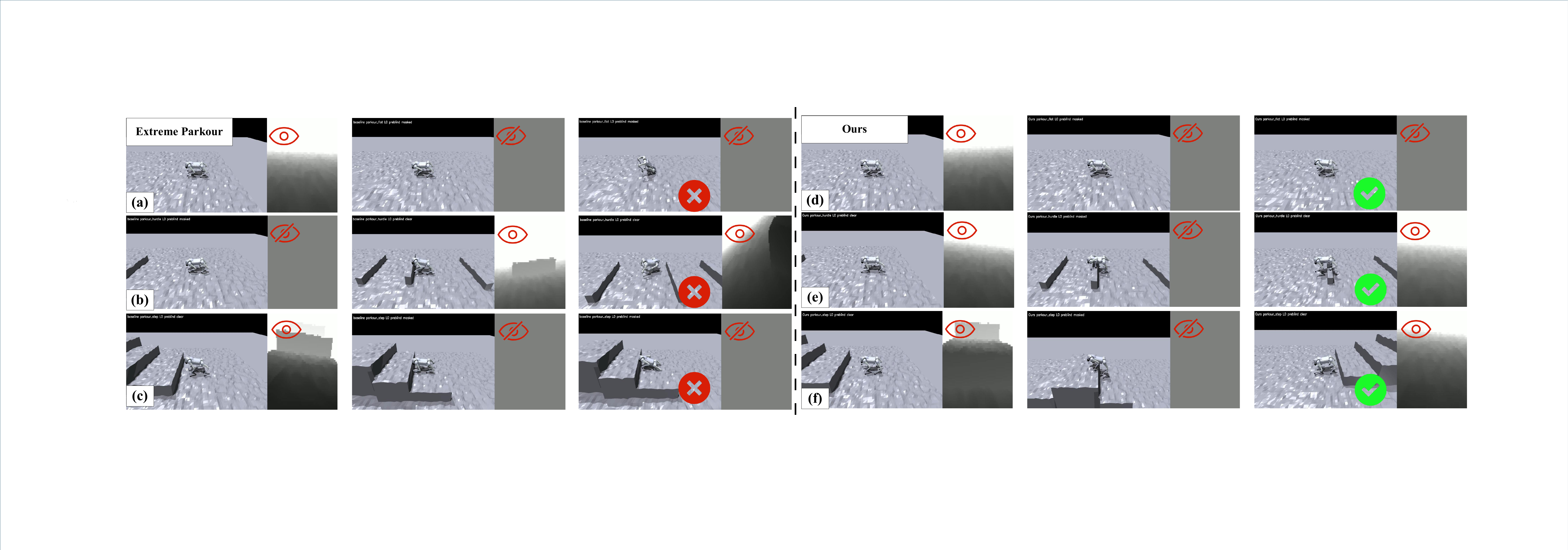}
    \caption{Simulation results on complex parkour terrains featuring a 1 m vision-masked blind zone (setup detailed in Table \ref{tab:memory_blind_zone}). REAL successfully traverses all terrains while maintaining kinematic stability.}
    \label{fig:main_exp}
\end{figure*}

% --- 域随机化参数表 最终修复版 --- 
\begin{table}[t]
\centering
\vspace{-2pt}
\caption{Domain Randomization Parameters for Go2} 
\label{tab:domain_randomization_go2}
\small % IEEE双栏规范表格字号
\setlength{\tabcolsep}{5pt} % 微调列间距，平衡留白与列宽
% 列宽权重分配：3列总权重必须=3，第一列占1.5倍宽度，后两列各占0.75倍
\begin{tabularx}{\linewidth}{
    >{\RaggedRight\arraybackslash\hsize=1.5\hsize}X  % 第一列：更多宽度+自动换行
    >{\centering\arraybackslash\hsize=0.75\hsize}X   % 第二列：居中+固定权重
    >{\centering\arraybackslash\hsize=0.75\hsize}X   % 第三列：居中+固定权重
} 
\toprule 
\textbf{Parameter} & \textbf{Range / Value} & \textbf{Distribution} \\ 
\midrule 
\multicolumn{3}{c}{\textit{Environment Dynamics}} \\ 
\midrule 
Terrain Friction & $[0.6, 2.0]$ & Uniform \\ 
External Push Interval & $8$ s & Fixed \\ 
% \mbox{}强制括号和前文在同一行，禁止拆行
External Push Velocity \mbox{(XY)} & $[-0.5, 0.5]$ m/s & Uniform \\ 
\midrule 
\multicolumn{3}{c}{\textit{Robot Kinematics \& Dynamics}} \\ 
\midrule 
Payload Mass \mbox{(Base)} & $[0.0, +3.0]$ kg & Uniform \\ 
Center of Mass \mbox{(CoM)} Offset & $[-0.2, 0.2]$ m & Uniform \\ 
\midrule 
\multicolumn{3}{c}{\textit{Actuator \& Observation Deficits}} \\ 
\midrule 
Motor Strength \mbox{(Kp, Kd Scaling)} & $[0.8, 1.2]$ & Uniform \\ 
Proprioceptive Noise \mbox{(Joint Pos)} & $\pm 0.01$ rad & Uniform \\ 
Proprioceptive Noise \mbox{(Joint Vel)} & $\pm 0.05$ rad/s & Uniform \\ 
Base Linear Velocity Noise & $\pm 0.05$ m/s & Uniform \\ 
Base Angular Velocity Noise & $\pm 0.05$ rad/s & Uniform \\ 
Gravity Vector Noise & $\pm 0.02$ & Uniform \\ 
Height Measurement Noise & $\pm 0.02$ m & Uniform \\ 
\bottomrule 
\end{tabularx}
\end{table}
\noindent\textbf{Domain Randomization.} To bridge the reality gap, we employ extensive domain randomization during training (Table \ref{tab:domain_randomization_go2}). We introduce perturbations to environmental dynamics (friction and mass), robot kinematics (center of mass and joint damping), and sensory observations (actuator latency and Gaussian noise). This diverse sampling compels the Mamba and FiLM modules to prioritize noise-invariant latent features, facilitating zero-shot transfer to the real world without further fine-tuning.

\noindent\textbf{Reward Formulation.} To ensure performance gains are strictly attributable to our architectural design rather than reward engineering, we adopt the exact reward formulation and weights utilized in Extreme Parkour. \cite{cheng2024extreme}.

\noindent\textbf{Baselines \& Metrics.} We compare REAL with Extreme-Parkour \cite{cheng2024extreme}, RPL \cite{zhuang2023robot}, and SoloParkour \cite{chane2024soloparkour}. As detailed in Table \ref{tab:extreme_terrain}, we adopt Collision Rate, Mean X-Displacement (MXD), and Mean Edge Violations (MEV) from \cite{cheng2024extreme}, while using custom definitions for Success Rate (SR) and Timeout Rate.

% --- B. 仿真主实验 ---
% \vspace{-2pt}
\subsection{Main Experiments}
\noindent\textbf{Extreme Terrain Traversability.} We adopt the exact extreme terrain configurations and difficulty scaling defined in \cite{cheng2024extreme}. As reported in Table \ref{tab:extreme_terrain}, REAL achieves a high overall success rate across these environments, effectively doubling the performance of prior vision-only baselines. Specifically, REAL exhibits superior precision when navigating hurdles and steps, where accurate foot placement is critical. While baseline methods suffer from high Mean Edge Violations (MEV) due to poorly planned maneuvers, our approach maintains a significantly lower MEV by relying on the robust spatio-temporal terrain memory built during deployment.

\begin{figure}[t]
    \centering
    \includegraphics[width=\columnwidth]{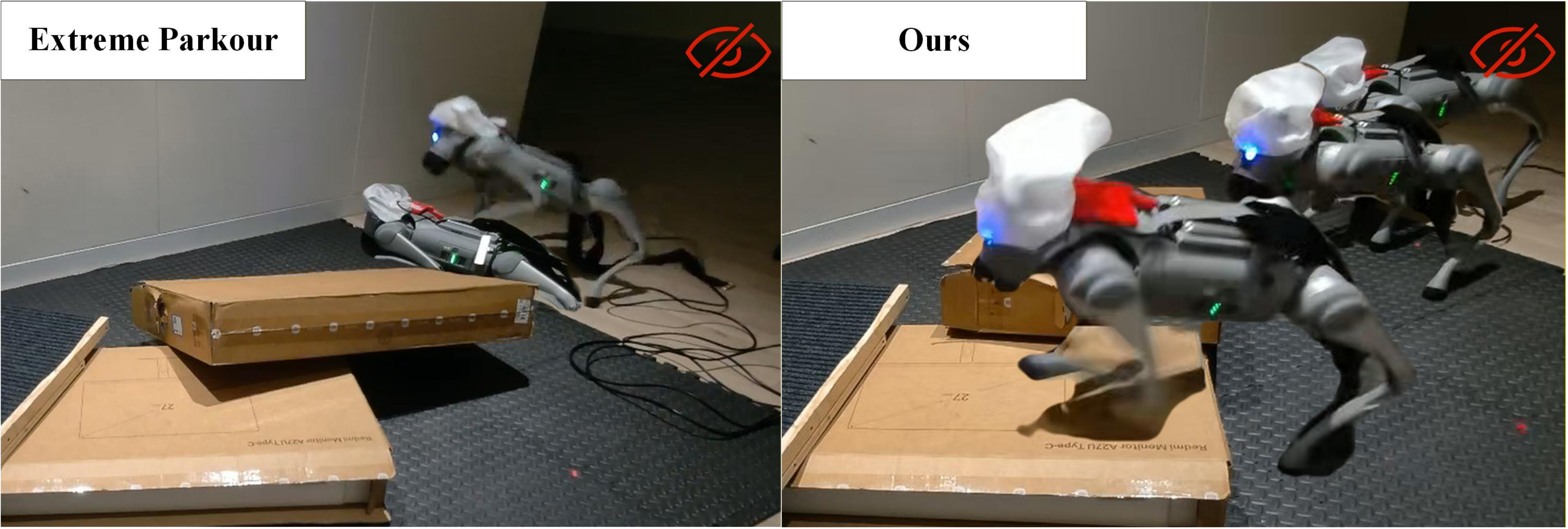}
    \caption{Real-world extreme blind test. \textbf{Left:} Baseline fails immediately upon losing visual input. \textbf{Right:} REAL utilizes proprioceptive history to maintain environmental memory, enabling robust blind traversal across unstructured obstacles.}
    \label{fig:realbot}
\end{figure}
\begin{table}[t]
\centering
\vspace{-2pt}
\caption{Quantitative Performance in Blind-Zone Maneuvers (Vision Masked 1m Before Obstacles)}
\label{tab:memory_blind_zone}
\resizebox{\columnwidth}{!}{%
\begin{tabular}{lccccc}
\toprule
\textbf{Method} & \textbf{SR ($\uparrow$)} & \textbf{MXD ($\uparrow$)} & \textbf{MEV ($\downarrow$)} & \textbf{Time ($\downarrow$)} & \textbf{Coll. ($\downarrow$)} \\
\midrule
Extreme Parkour & 0.11 & 0.20 & 44.03 & 0.15 & 0.29 \\
RPL             & 0.00 & 0.03 & 0.35  & 0.00 & 0.04 \\
SoloParkour     & 0.36 & 0.34 & 103.50 & 0.06 & 0.09 \\
\midrule
\textbf{REAL (Ours)} & \textbf{0.55} & \textbf{0.39} & \textbf{24.84} & \textbf{0.03} & \textbf{0.08} \\
REAL (w/o Mamba)& 0.45 & 0.51 & 97.24 & 0.30 & 0.06 \\
\bottomrule
\end{tabular}%
}
\end{table}

\noindent\textbf{Robustness Against Perceptual Degradation.} Table \ref{tab:robustness_drop} evaluates policy performance under simulated sensor degradation, including dropped frames, Gaussian noise, and obstructed field of view (FoV). Under severe FoV occlusion, REAL experiences only a marginal performance decline. This robustness highlights the efficacy of our Cross-Modal mechanism, which seamlessly relies on proprioceptive history when visual data is compromised. Conversely, vision-reliant multi-stage baselines suffer catastrophic failures in the presence of sensor noise, underscoring the vulnerability of methods lacking robust spatio-temporal grounding.

\noindent\textbf{Blind-Zone Maneuvers.} We evaluate spatio-temporal memory by obstructing exteroception 1 m before obstacles (Table \ref{tab:memory_blind_zone}, Fig. \ref{fig:realbot}). Without vision, standard baselines suffer catastrophic forgetting and immediate kinematic failure. Conversely, REAL leverages historical multi-modal data to implicitly track the terrain. Although perfectly clearing complex obstacles blind remains an out-of-distribution challenge causing unstable movements, REAL maintains significantly stronger forward progression than baselines. We exclusively ablate vision, as complete loss of proprioception triggers immediate low-level motor halts.

\begin{figure}[htbp]
    \centering
    \vspace{5pt}
\includegraphics[width=0.9\columnwidth]{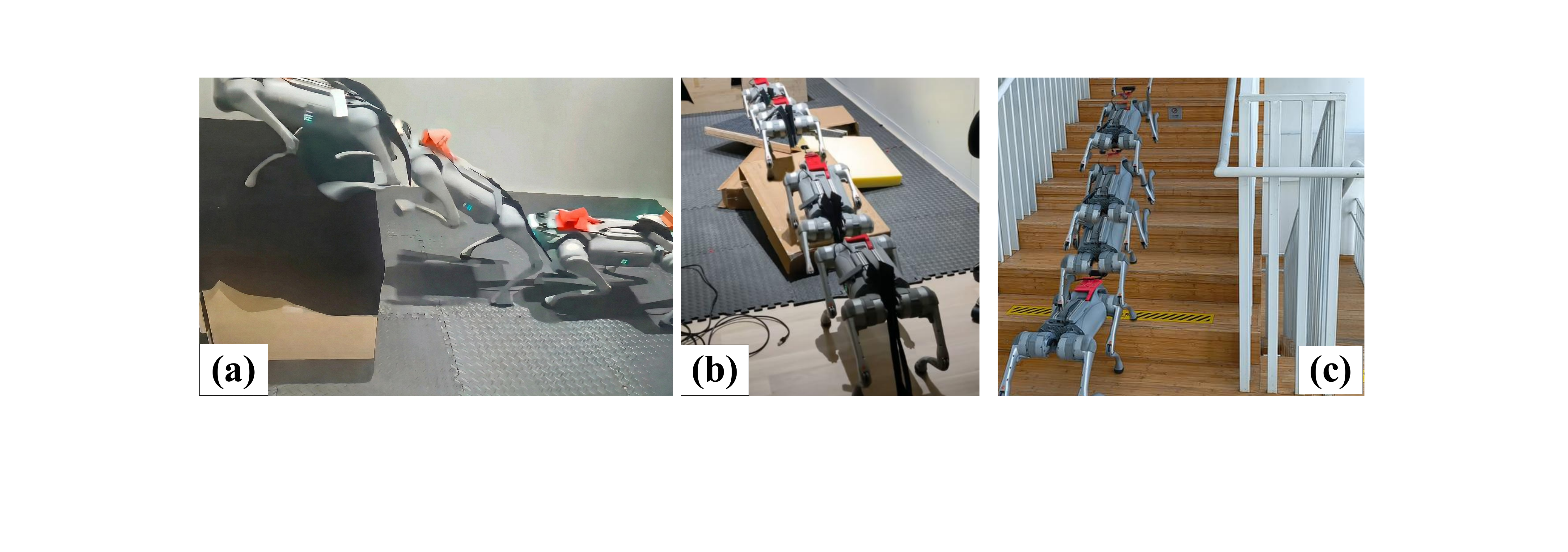}
    \caption{Zero-shot sim-to-real transfer of the REAL policy on a physical Unitree Go2 quadruped. Using only onboard perception and computing, the robot completes various real-world obstacle courses: \textbf{(a)} leaping onto a high platform, \textbf{(b)} moving through scattered boxes, and \textbf{(c)} climbing a steep staircase.}
    \label{fig:firstpage}
\end{figure}

\begin{figure}[htbp]
    \centering
    \includegraphics[width=0.95\columnwidth]{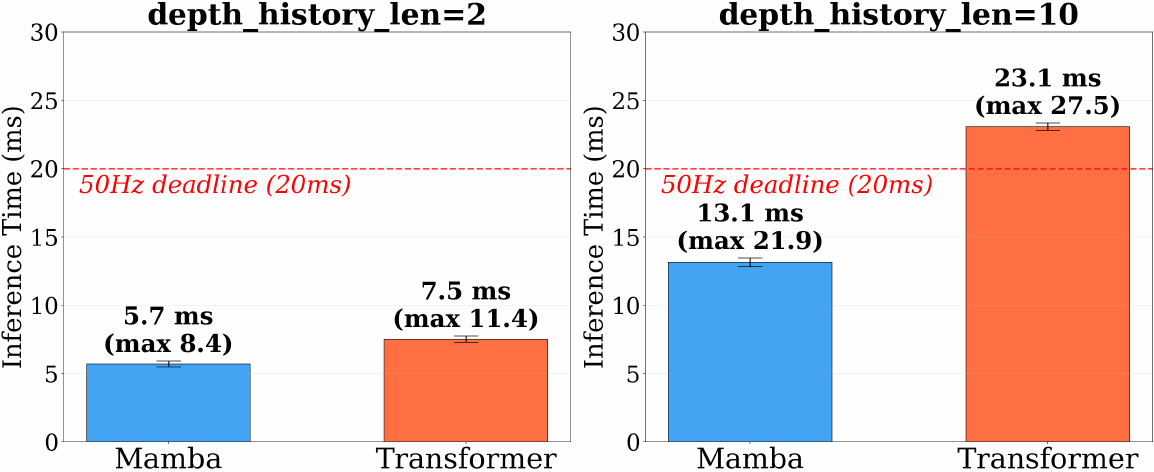}
    \caption{Onboard inference latency measured over 1,000 continuous control steps. The REAL policy maintains a highly predictable execution time of $O(1)$ ($\sim 13.1$ ms/step), strictly satisfying the high-frequency real-time control budget.}
    \label{fig:inference_time}
\end{figure}
% --- C. 实际机器人实验 ---
\subsection{Real-Robot Deployment}
%\noindent\textbf{Onboard Real-Time Execution.} 
As demonstrated in Fig. \ref{fig:firstpage}, our trained REAL policy achieves robust zero-shot sim-to-real transfer on a physical Unitree Go2 quadruped, completing diverse real-world obstacle courses with only onboard perception and computing. We deploy the policy onto the Go2 hardware via a custom C++ framework, optimizing both the policy and state estimator with ONNX. To maintain a stable 50 Hz control loop, the total inference time must stay strictly under 20 ms. As illustrated in Fig. \ref{fig:inference_time}, we evaluate inference latency across different temporal sequence lengths. With a 10-frame historical sequence, our Mamba backbone achieves an average execution time of 13.14 ms per step. In stark contrast, a Transformer-based baseline scales poorly, averaging 23.07 ms and violating the 20 ms real-time constraint. This result confirms that Mamba's bounded $O(1)$ complexity eliminates the sequence-scaling bottleneck of Transformers, enabling the high-frequency reactivity required for aggressive parkour maneuvers.

\subsection{Ablation Study}
\begin{table}[t]
\centering
\vspace{-2pt}
\caption{Overall Ablation Analysis of REAL Components}
\label{tab:ablation_overall}
% 使用 resizebox 确保表格不会超出边距
\resizebox{\columnwidth}{!}{
\begin{tabular}{l c c c c c}
\toprule
\textbf{Method} & \textbf{SR ($\uparrow$)} & \textbf{MXD ($\uparrow$)} & \textbf{MEV ($\downarrow$)} & \textbf{Time ($\downarrow$)} & \textbf{Coll. ($\downarrow$)} \\
\midrule
\textbf{REAL (Ours)} & \textbf{0.78} & 0.45 & \textbf{18.41} & \textbf{0.02} & 0.06 \\
REAL (w/ MLP Est.) & 0.73 & 0.43 & 19.34 & \textbf{0.02} & 0.06 \\
REAL (w/o FiLM) & 0.44 & \textbf{0.51} & 93.43 & 0.28 & 0.06 \\
REAL (w/o Mamba) & 0.51 & 0.47 & 89.96 & 0.26 & \textbf{0.05} \\
\bottomrule
\end{tabular}
}
% \vspace{1mm}
\begin{flushleft}
% \footnotesize{Note: \textbf{SR}: Success Rate, \textbf{MXD}: Mean X-Displacement, \textbf{MEV}: Mean Edge Violation, \textbf{Coll.}: Collision Rate. Best results are highlighted in \textbf{bold}.}
\end{flushleft}
\end{table}
\begin{table}[h!]
\centering
\vspace{-2pt}
\caption{Velocity Estimation RMSE Comparison}
\label{tab:estimator_rmse}
\small % IEEE标准：表格字号比正文小一号，同时压缩横向宽度
\setlength{\tabcolsep}{4pt} % 微调列间距，减少横向占用（默认6pt）
\begin{tabular}{lc}
\toprule
\textbf{Estimator Architecture} & \textbf{RMSE (Velocity) $\downarrow$} \\
\midrule
MLP (Baseline) & 0.52 \\
MLP + EKF & 0.40 \\
1D ResNet (Single frame) & 0.33 \\
1D ResNet (10 frames) & 0.28\\
\textbf{1D ResNet + EKF (10 frames, Ours)} & \textbf{0.23} \\
\bottomrule
\end{tabular}
\end{table}
\begin{figure}[htbp]
    \centering
    \includegraphics[width=0.8\columnwidth]{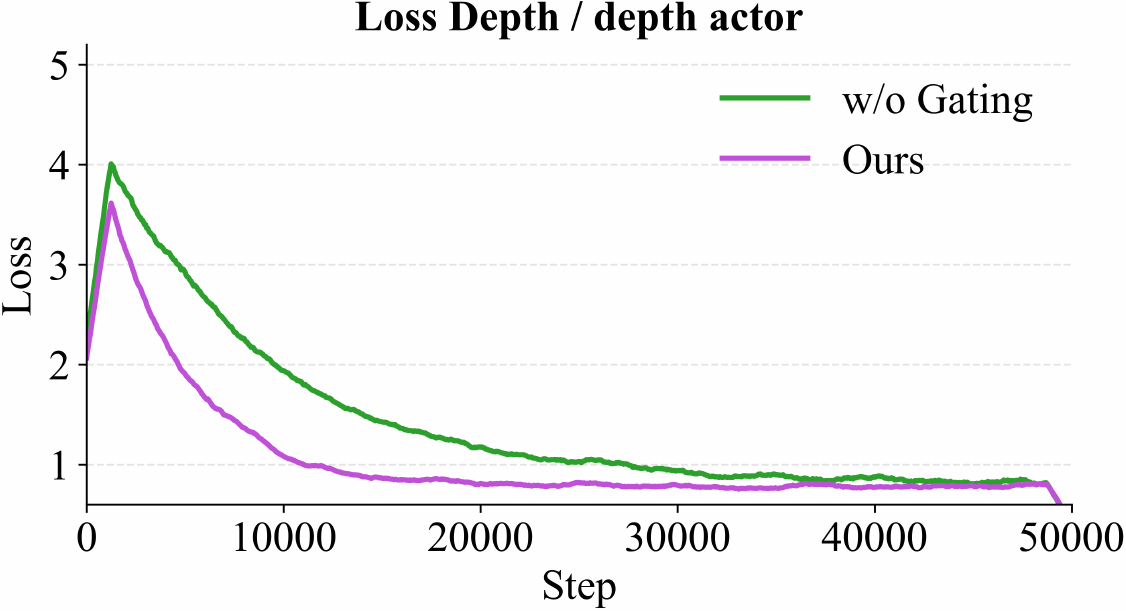} 
    \caption{Training convergence of the depth actor. Our consistency-aware loss gating accelerates imitation learning and achieves a lower final loss compared to a fixed-weight baseline, improving the stability of the distillation process.}
    \label{fig:gating_ablation}
\end{figure}
\noindent\textbf{The Role of Mamba and FiLM.} We perform ablation studies to isolate the contribution of our core architectural components in maintaining kinematic stability (Table \ref{tab:ablation_overall}).
\begin{itemize}
    \item \textit{W/o Mamba}: Removing the Mamba backbone causes the SR to plummet and MEV to increase nearly fivefold (from 18.41 to 89.96). This confirms that Mamba's long-term sequence modeling is strictly necessary for tracking historical terrain features when exteroception is lost.
    \item \textit{W/o FiLM}: Disabling the dynamic sensory gating mechanism drops the success rate to 44\%. Without FiLM, the policy fails to robustly fuse multi-modal signals, treating noisy visual inputs as ground truth and leading to frequent torso collisions.
\end{itemize}
% \vspace{-2pt}
\noindent\textbf{Physics-Guided Filtering Validation.} As demonstrated in Table \ref{tab:estimator_rmse}, standard MLP-based estimators suffer from high velocity tracking errors due to a lack of temporal context. Transitioning to a sequence-based ResNet model with a 10-frame proprioceptive sequence drastically reduces these errors. Our Physics-Guided Filter achieves the lowest overall root mean square error (RMSE) by filtering high-frequency impact noise and enforcing strict physical constraints during leg-ground contact.

\noindent\textbf{Consistency-Aware Loss Gating.} To prevent policy collapse under aggressive domain randomization during the imitation phase, we introduce a consistency-aware loss gating strategy. Unlike fixed-weight baselines that frequently collapse during initial training, our dynamic gating optimally balances autonomous exploration with teacher imitation. As shown in Fig. \ref{fig:gating_ablation}, this mechanism drastically accelerates early-stage convergence and yields a significantly lower final training loss, providing the robust foundation required for reliable real-world locomotion.

\section{Conclusion}
In this paper, We presented REAL, an end-to-end framework for robust quadrupedal parkour under severe perceptual degradation. By integrating a FiLM-modulated Mamba backbone, a physics-guided Bayesian state estimator, and consistency-aware loss gating, REAL effectively constructs short-term spatio-temporal memory to suppress unreliable perception and ensure stable sim-to-real distillation. Hardware experiments on the Unitree Go2 demonstrate successful zero-shot transfer, outperforming state-of-the-art baselines across extreme terrains. Crucially, REAL exhibits exceptional resilience to sensory noise and blind zones, all while strictly satisfying the high-frequency real-time control budget with a predictable ${O}(1)$ inference latency.

% \clearpage
\bibliographystyle{ieeetr}
\bibliography{papers}

\clearpage

\end{document}